\documentclass[letterpaper, 10 pt, conference]{ieeeconf}  

\IEEEoverridecommandlockouts                              

\usepackage{amsmath} 
\usepackage{amssymb}  

\newcommand{\vv}{\bm{v}}

\usepackage{mathtools}
\DeclarePairedDelimiterX{\infdivx}[2]{[}{]}{%
  #1\;\delimsize\|\;#2%
}
\newcommand{\infdiv}{\text{KL}\infdivx}

\def\vv0i{\mathbf{v}_i^d}

\def\f0i{\mathbf{f}_i^0}

\newcommand{\FriW}[1]{\emph{FriWalk}}

\makeatletter
\def\squiggly{\bgroup \markoverwith{\textcolor{red}{\lower3.5\p@\hbox{\sixly \char58}}}\ULon}
\makeatother

\usepackage{bm}
\usepackage{pifont}
\usepackage{booktabs}
\usepackage{mathtools}
\newcommand{\cmark}{\ding{51}}
\newcommand{\xmark}{\ding{55}}
\newcommand{\algname}{\textbf{GEM-MPC }}
\newcommand{\algnamep}{\textbf{GEM-MPC}}
\newcommand{\githublink}{https://anonymous.4open.science/r/GEM-MPC-1E79}
\title{\LARGE \bf
GEM-MPC: Balancing Exploration and Exploitation through Expert-Guided Planning
}

\author{\'Alvaro Serra-Gomez and Thomas Moerland
\thanks{Both authors are with the Leiden Institute of Advanced Computer Science, Leiden University, 2311 EZ Leiden, The Netherlands.
        {\tt\small \{a.serra-gomez,t.m.moerland\}@liacs.leidenuniv.} {\tt\small nl}}%
}

\begin{document}

\maketitle
\thispagestyle{empty}
\pagestyle{empty}

\begin{abstract}

Effective exploration in high-dimensional continuous control remains a central challenge in reinforcement learning. Planning-based methods address this by combining online planning with learned policies and value functions, but their components can become misaligned during training: learned sampling policies may diverge from planner behavior, while planning distributions stored in replay become stale as the model and value function evolve. Reanalysis can refresh these targets, but at substantial computational cost.
We propose GEM-MPC, an MPPI-based reinforcement learning method that improves the interaction between planning and learning. GEM-MPC uses MPPI to combine a policy trained to clone the planner with a KL-regularized policy that explores around it, providing complementary exploitation and guided exploration within planning. We further introduce Gated Prior Distillation, which selectively learns from stored planning distributions only when they provide a better target than the current prior, reducing the impact of stale planning data without requiring full reanalysis.
Across continuous-control benchmarks, GEM-MPC consistently outperforms existing planning-based baselines under lower computational budgets.

\end{abstract}

\section{Introduction}
\label{sec:intro}
Exploration in high-dimensional continuous action spaces remains a major challenge in reinforcement learning. Planning-based methods address this challenge by combining shallow online search, such as Model Predictive Path Integral control (MPPI)~\cite{it-mppi}, with learned sampling policies and bootstrap value functions~\cite{tdmpc1,hansen2024tdmpc2}. The sampling policy, $\pi_{\theta_s}$, directs the planner toward promising regions of the action space, while the action-value function, $Q_{\theta_Q}^{\pi_{\theta_s}}$, extends the effective planning horizon.
 
However, learning the sampling policy independently from the planner creates a mismatch between $\pi_{\theta_s}$ and the MPPI-induced planning distribution $\pi^P$. This mismatch degrades value function estimation, reducing sampling efficiency and limiting final performance. Recent methods address it by imitating the planner distribution~\cite{wang2025bootstrapped} or using it as a prior for policy regularization~\cite{lin2025td,zhan2025bootstrap}. 

PO-MPC~\cite{serragomez2026pompc} further interprets this interaction as KL-regularized reinforcement learning, where the sampling policy explores high-value actions around a policy prior, $\pi_{\theta_p}$, that acts as a proxy for the planner distribution. This formulation provides effective guided exploration, but does not fully resolve exploitation. The sampling policy may concentrate on only a subset of the local maxima of the action-value function around the learned prior, without necessarily preserving useful modes represented by the learned prior itself. It is therefore well suited for exploration around the planner, while stable exploitation depends on maintaining a high-quality planning samples.
 
Learning a reliable planner representation heavily depends on the planning distributions stored in the replay buffer. Unlike transition data, which retains valid information about the environment's transition function, planning data depends on the model and value function at the time of generation. As these get updated throughout training, stored planning targets may become stale and introduce noisy or harmful behavior.
 \begin{figure}[t]
\centering
\includegraphics[width=0.49\textwidth]{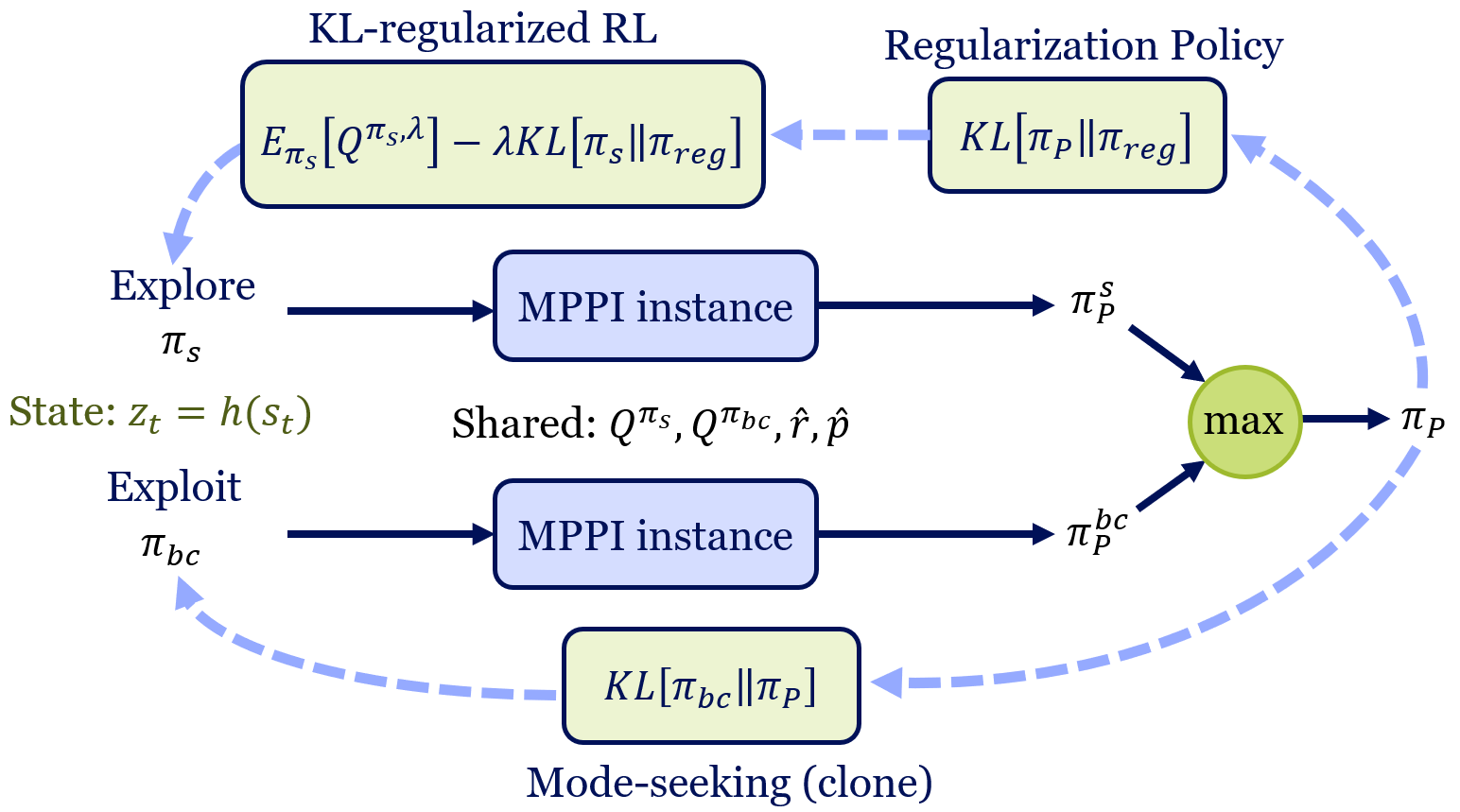}
\caption{\textbf{Overview of GEM-MPC.} Two independent MPPI instances compute distinct planning distributions, from which the higher-scoring plan is selected according to its estimated return. Both instances share the learned environment model and are biased by the sampling policies $\pi_s$ and $\pi_{bc}$ together with their corresponding bootstrap action-value functions. The selected planning distribution is then used to update the learned sampling policies. The regularized sampling policy $\pi_s$ is trained with KL-regularized RL, using a regularization policy learned with a support-covering objective over stored planning distributions, while the cloned policy $\pi_{bc}$ is trained with a mode-seeking behavior-cloning objective that concentrates on high-density planner behavior.}
\label{fig:overview}
\end{figure}
Reanalyze and Lazy reanalyze methods~\cite{schrittwieser2021online, wang2025bootstrapped} address this problem by recomputing planning solutions for stored environment transitions. However, such methods are computationally expensive and do not necessarily solve the problem, since a newly recomputed target is not necessarily better: an inaccurate or overconfident value function may replace a useful high-entropy historical target with a planning distribution that commits to an incorrect region of the action space.

This work proposes \textbf{Gated distillation with Expert Mixtures for Model Predictive Control} (\algnamep), an MPPI-based RL method that automatically balances exploration and exploitation while providing an alternative to Reanalyzing strategies. Building upon PO-MPC~\cite{serragomez2026pompc}, this paper uses MPPI to seamlessly combine two policies: a greedy policy that clones planner, and an exploratory policy that explores around it. The former is trained through Behavior Cloning (BC), while the latter is trained through KL-regularized RL. To avoid harmful behavior being distilled from stale planning distributions, we introduce a simple gating mechanism that distills a stored planning distribution only when it provides a better target than the current prior without requiring full reanalysis for every replayed state~\cite{wang2025bootstrapped}. In summary, our contributions are:
\begin{itemize}
     \item \textbf{Expert-augmented planning:} a mixture-of-experts formulation that combines both a cloned policy, and an exploratory policy learned through KL-regularized RL to balance exploration and exploitation within planning.
    \item \textbf{Gated Prior Distillation:} a theoretically motivated gating mechanism that filters stale or harmful planning targets, resulting in planning distribution representation at substantially lower computational cost than reanalysis.
\end{itemize}

 
 We show that the proposed method consistently improves existing planning-based reinforcement learning baselines~\cite{serragomez2026pompc,wang2025bootstrapped} under lower computational budgets. An overview of our method's structure is presented in Fig.~\ref{fig:overview}. 

\section{Related Work}
\label{sec:related}

Model-based reinforcement learning combines environment models with policy and value learning to improve sequential decision-making~\cite{moerland2023model}. Our work focuses on methods that use online planning, i.e. MPPI~\cite{it-mppi}, both to select actions and to guide learning.\\

\textbf{MPPI-based MBRL methods} exploit a reciprocal relationship between planning and learning. In TD-MPC and TD-MPC2, learned policies provide trajectory proposals, while value functions estimate returns beyond the planning horizon~\cite{tdmpc1,hansen2024tdmpc2}. Conversely, planning can provide supervision for policy learning. BMPC learns a policy by imitating the planner's action  distribution~\cite{wang2025bootstrapped}, while PO-MPC uses a learned representation of this distribution as a prior for KL-regularized policy optimization~\cite{serragomez2026pompc}. Similar approaches follow similar guidelines underscoring the importance of policy learning and planning alignment~\cite{lin2025td} to avoid sampling out-of-distribution state-action couples.\\



\textbf{Exploration and exploitation during planning.} Sampling-based planning must concentrate its search on promising actions while retaining sufficient diversity to discover alternatives. Entropy regularization and learned behavior priors offer mechanisms for shaping policy exploration~\cite{haarnoja2018soft,tirumala2022}. Other approaches modify trajectory evaluation through uncertainty penalties~\cite{evers2026efficienttdmpc}, pessimistic value estimates to avoid overestimation in out-of-distribution state-action couples~\cite{chang2026the}, or maintain independent trajectory optimizers to reduce susceptibility to local optima, as in DecentCEM~\cite{DecentCEMzhang2022a}. Building on PO-MPC, our method combines a planner-cloned policy for exploitation with a KL-regularized policy for guided exploration. Separate MPPI branches preserve their complementary sampling biases during planning.\\


\textbf{Learning from stored planning results.} Planning distributions stored in replay can become outdated as models, policies, and value functions evolve. Reanalysis addresses this problem by recomputing planning targets for previously collected experience, as in MuZero Reanalyse and BMPC's lazy reanalyze~\cite{schrittwieser2021online,wang2025bootstrapped}. A related line of work accounts for differences in stored target quality through value-based weighting of planner alignment using the current Q-function~\cite{zhan2025bootstrap,zhuang2025tdmpbc}. Our work follows an alternative path, similar to demonstration Q-filters in Model-free RL, which apply imitation only when a demonstrated action is judged better than the policy's action~\cite{nair2018overcoming, CRR_NEURIPS2020_588cb956}. Gated Prior Distillation follows this selective-imitation perspective, using value-based comparisons to determine which stored planner distributions are not detrimental when learning the planning distribution. This allows planning experience to be reused selectively without recomputing each target. 

\section{Preliminaries}\label{sec:preliminaries}

We consider a discrete-time sequential decision-making problem over a finite horizon of \(T\) steps, modeled as a Markov decision process (MDP) \((\mathcal{S},\mathcal{A},p,r,\gamma)\). Here, \(\mathcal{S}\) and \(\mathcal{A}\) denote the state and action spaces, respectively, \(p(\cdot\mid s,a)\) is the transition distribution, encompassing both stochastic and deterministic dynamics, \(r(s,a)\) is the immediate reward for taking action \(a\) in state \(s\), and \(\gamma\in[0,1)\) is the discount factor. A policy \(\pi(\cdot\mid s)\) specifies a distribution over actions in state \(s\). The objective is to find a policy that maximizes the expected discounted return:

$$
J(\pi)
=
\mathbb{E}_{\substack{
s_0\sim\rho_0,\; a_t\sim\pi(\cdot\mid s_t),\\
s_{t+1}\sim p(\cdot\mid s_t,a_t)
}}
\left[\sum_{t=0}^{T-1}\gamma^t r(s_t,a_t)\right],
$$

where \(\rho_0\) denotes the initial state distribution.

\textbf{MPPI-based Reinforcement Learning.}
Recent methods in model-based reinforcement learning integrate online planning directly into the learning loop, with approaches based on Model Predictive Path Integral Control (MPPI)~\cite{it-mppi} proving particularly effective in high-dimensional robotic control~\cite{tdmpc1,hansen2024tdmpc2}. These methods learn a latent model of the environment,
$(\hat{\mathcal{S}},\mathcal{A},\hat{p},\hat{r},\gamma)$,
where observations or states are mapped to a latent representation
$z=h_{\theta_h}(s)\in\hat{\mathcal S}$, and planning is performed using learned reward and transition models
$\hat r(z,a)=r_{\theta_r}(z,a)$ and
$\hat p=p_{\theta_d}$~\cite{bhardwaj2021blending}.

MPPI is a sample-based stochastic planning method that iteratively refines a distribution over finite-horizon action sequences according to their predicted returns. Let $H$ denote the planning horizon, and
$\bar a_{0:H-1}=(\bar a_0,\ldots,\bar a_{H-1})$
the mean of the current open-loop action distribution. At each planning iteration, MPPI samples $M$ perturbed action sequences,
\begin{equation}
a_t^{(i)}
=
\bar a_t+\epsilon_t^{(i)},
\qquad
\epsilon_t^{(i)}
\sim
\mathcal{N}(0,\sigma_t^2 I),
\end{equation}
and rolls them out through the learned latent dynamics,
\begin{equation}
z_{t+1}^{(i)}
\sim
p_{\theta_d}
\left(
\cdot\mid z_t^{(i)},a_t^{(i)}
\right).
\end{equation}
The resulting trajectories $\tau_i$ are evaluated using their predicted returns:
\begin{equation}
R(\tau_i)
=
\sum_{t=0}^{H-1}
\hat{r}\left(z_t^{(i)},a_t^{(i)}\right).
\end{equation}
MPPI assigns exponentially larger weight to trajectories with higher predicted returns,
\begin{equation}
w_i
=
\frac{
\exp\left(R(\tau_i)/\lambda\right)
}{
\sum_j
\exp\left(R(\tau_j)/\lambda\right)
},
\end{equation}
where $\lambda>0$ controls the concentration of the weighting distribution. The parameters of the action distribution are then updated using the weighted perturbations,
\begin{align}\label{eq:mppi_update}
\bar a_t
&\leftarrow
\bar a_t+\sum_{i=1}^{K}w_i\epsilon_t^{(i)},\
\sigma_t
&\leftarrow
\sqrt{
\frac{
\sum_{i=1}^{K}
w_i(\epsilon_t^{(i)})^2
}{
\sum_{i=1}^{K}w_i
}
},
\end{align}
where the update may be restricted to the $K$ highest-return trajectories. After a fixed number of iterations, the optimized distribution over the first action defines the \textbf{planning policy},
\begin{equation}
\pi_{\mathrm P}(\cdot\mid z)
=
\mathcal N(\bar a_0,\sigma_0^2 I).
\end{equation}
Only the first action is executed before replanning from the next state, yielding a receding-horizon feedback policy. The mean sequence can additionally be warm-started by shifting the solution from the previous decision step by one time step.

While standard MPPI samples from a broad Gaussian proposal, recent model-based RL methods additionally generate part of the candidate trajectories using a \textbf{learned sampling policy} $\pi_{\theta_s}$~\cite{hansen2024tdmpc2,wang2025bootstrapped}. This biases the finite planning budget toward action sequences that are already expected to achieve high return, while the higher-variance MPPI proposal preserves exploration outside the current learned policy.

The learned sampling policy is also used to estimate returns beyond the finite planning horizon. In particular, an action-value function
$Q_{\theta_Q}^{\pi_{\theta_s}}$
bootstraps the terminal latent state of each rollout, yielding the $H$-step estimate
\begin{equation}\label{eq:mppi_evaluation}
Q(z_0,a_{0:H}^{(i)})
=
\sum_{t=0}^{H-1}
\gamma^t
r_{\theta_r}
\left(
z_t^{(i)},a_t^{(i)}
\right)
+
\gamma^H
Q_{\theta_Q}^{\pi_{\theta_s}}
\left(
z_H^{(i)},a_H^{(i)}
\right).
\end{equation}
Thus, planning operates over a mixture of trajectories proposed by the learned sampling policy and by the broader MPPI distribution. The learned policy concentrates samples in regions of the action space that are already predicted to be useful, whereas the MPPI proposal retains broader coverage. MPPI subsequently reweights these candidate trajectories according to their predicted $H$-step returns, combining policy-guided sampling with online trajectory optimization.

The method proposed by this paper is built PO-MPC~\cite{serragomez2026pompc}, which unifies a subset of MPPI-based algorithms~\cite{hansen2024tdmpc2,wang2025bootstrapped} by framing sampling policy learning as KL-regularized RL. Under this framework, the sampling policy is trained to maximize its action-value function while being regularized to generate trajectories that lie within the planner's trajectory distribution. For that reason, a learned policy that approximates the planner is used as regularizer. The sampling policy is updated by maximizing the following objective function:
\begin{align}\label{eq:pompc_policy_update}
J(\pi)=&\mathbb{E}_{a\sim \pi_{\theta_s}}[Q^{\pi_{\theta_s},\lambda}_{\tilde\theta_Q}(z_t,a_t)] \\
&-\lambda \infdiv{\pi_{\theta_s}(\cdot\mid z_t)}{\pi_{reg}(\cdot\mid z_t)}, \notag
\end{align}

where the regularized Q-value function $Q^{\pi_{\theta_s},\lambda}_{\tilde\theta_Q}$ satisfies the following Bellman recursive expression:

\begin{align}\label{eq:pompc_regQ_update}
    Q^{\pi_{\theta_s},\lambda}_{\tilde\theta_Q}&(z_t,a_t) =\mathbb{E}_{\substack{s_{t+1}\sim p(\cdot\mid s_{t},a_{t}),\\a\sim\pi_{\theta_s}(\cdot\mid z_{t+1})}}\Biggl[r(z_t,a_t)\\ 
    &+\gamma\Biggl(Q^{\pi_{\theta_s},\lambda}_{\tilde\theta_Q}(z_{t+1},a) - \lambda\log\Biggl(\frac{\pi_{\theta_s}(a\mid z_{t+1})}{\pi_{reg}(a\mid z_{t+1})}\Biggr)\Biggr)\Biggr] \notag
\end{align}

Note that, as remarked in~\cite{serragomez2026pompc}, there are many ways to train the regularization policy. This results in different inductive biases being embedded into the sampling policy, resulting in faster convergence or enhanced exploration during planning. We shall leverage this property in the following section. As in PO-MPC, the rest of this work borrows world model learning from TD-MPC2, and maintains the architecture choices for the encoder, policies and action-value functions.

\section{Method}
\label{sec:method}


An overview of our method is presented in Fig.~\ref{fig:overview}. This work addresses two limitations of MPPI-based reinforcement learning: the exploration-exploitation trade-off during planning and the use of stale planning distributions for learning sampling policies.


Prior approaches typically bootstrap MPPI with a single sampling policy and its corresponding action-value function, thereby biasing planning toward either exploration or exploitation depending on how the sampling policy is trained. In high-dimensional, non-convex action-value landscapes, this can lead the planner to overcommit to local optima or remain around more stable but suboptimal solutions. We instead combine exploratory and exploitative sampling policies, together with their corresponding action-value functions, to provide complementary proposals during planning.


However, this requires access to the current planning policy. A separate challenge arises when distilling the resulting planning distributions into learned sampling policies. Ideally, these policies would be trained against the current planner, but obtaining up-to-date planning targets requires re-planning from replayed states. Prior methods either learn directly from stored planning distributions, which become stale as the policies and value functions evolve, or mitigate this mismatch through reanalysis, i.e., re-planning transitions from the replay buffer at substantial computational cost. We instead introduce a theoretically motivated gating criterion that selectively distills stored planning targets according to their action-value under the current learned policies, providing a lower-cost alternative to reanalysis.


\subsection{Expert-augmented Planning via Decentralized MPPI}\label{sec:MoE}

\textbf{Biased Trajectory Sampling.} 
Candidate trajectories are generated from a three-component sampling scheme composed of the planning distribution $\pi_P$ and two biasing policies $\pi_{\theta_{s}}$ and $\pi_{\theta_{bc}}$. If a candidate is sampled from $\pi_P$ with probability $(1-p)$ and from $\pi_i,~i\in\{\theta_s,\theta_{bc}\}$ each with equal probability $\frac{p}{2}$, the induced trajectory distribution is $\hat\pi_s(\tau) = (1-p)\pi_P(\tau) + \frac{p}{2}\pi_{\theta_{s}}(\tau)+\frac{p}{2}\pi_{\theta_{bc}}(\tau)$, where each component denotes the trajectory distribution induced by the corresponding action-sequence sampling mechanisms and the environment dynamics.

Assuming access to previously computed planning distributions stored in the replay buffer, the aim of the first policy, the regularized sampling policy: $\pi_{\theta_s}$, is to provide a distribution that maximizes the action value function while remaining close to the planner distribution. Therefore, it is trained using KL-regularized RL, as in PO-MPC~\cite{serragomez2026pompc}, where the policy updates are regularized
using a regularization policy, $\pi_{\theta_{reg}}$, that keeps $\pi_{\theta_s}$ trajectories close to the planner's  following Equation~\ref{eq:pompc_policy_update}.

While the regularization policy should be equal to the planner distribution, this one is constantly changing because it depends on the current state of the bootstrap action-value functions, which are also constantly being updated. For that reason, we use a learned regularization policy that clones the planning distributions. In order not to prematurely dismiss part of the support of stored MPPI samples, which might turn out to contain high local maxima of the action-value function, the regularization policy is trained to clone the stored planner samples while discouraging assigning low probability to regions with high planner density. 
This is done by minimizing the following forward KL divergence:

\begin{equation}\label{eq:fkl_loss}
J(\theta_{reg})=\mathbb{E}_{(s,\pi_P)\sim D}\Bigl[\infdiv{\pi_P(\cdot\mid z_t)}{\pi_{\theta_{reg}}(\cdot\mid z_t)}\Bigr],    
\end{equation}

where $D$ is the data stored in the replay buffer, and $\pi_P$ is the stored planning distribution.

On the contrary, the goal of the second policy, the cloned sampling policy $\pi_{\theta_{bc}}$, is to clone closely the most likely behavior of the planner. This is why it is trained to concentrate on high-density regions of the planning distributions stored within the replay buffer:
\begin{equation}\label{eq:rkl_loss}
J(\theta_{bc})=\mathbb{E}_{(s,\pi_P)\sim D}\Bigl[\infdiv{\pi_{\theta_{bc}}(\cdot\mid z_t)}{\pi_{P}(\cdot\mid z_t)}\Bigr].
\end{equation}

The application of both distillation objectives to potentially stale planning targets is governed by the gating mechanism introduced in Sec.~\ref{sec:GPD}.\\


\textbf{Trajectory evaluation.} Typically, simulated trajectories are evaluated using the modeled H-step action-value function, where $H$ is the planning horizon (Eq.~\ref{eq:mppi_evaluation}).

This implies using the modeled cumulative reward plus a bootstrap action value function of the learned policy is used to bias planning. Assuming the sampling policy is followed past the planning horizon, the bootstrap action-value function is an estimate of the expected value of the trajectory past the planning horizon. 

Instead, our method bootstraps each trajectory with the mean of the action-value function of each policy, $\pi_{\theta_s}$ and $\pi_{\theta_{bc}}$, evaluated in actions sampled from their respective distributions: 

\begin{align}\label{eq:alg_evaluation}
    Q(z_0,a_{0:H}^{(i)}) &= \sum^{H-1}_{t=0}\gamma^t r_{\theta_r}(z_t,a_t^{s,(i)}) \notag\\
    &\quad+\frac{\gamma^H}{2}(Q_{\theta_{Q_s}}^{\pi_{\theta_s}}(z_H,a_H^{(i)})+Q_{\theta_{Q_{bc}}}^{\pi_{\theta_{bc}}}(z_H,a_H^{bc,(i)})).
\end{align}

Where $a_t^{s,(i)}$ and $a_t^{s,(i)}$ are sampled according to $\pi_{\theta_s}$ and $\pi_{\theta_{bc}}$. Once the planning horizon limit is reached, this is equivalent, in expectation, to randomly choosing one out of the two policies and using it until reaching the end of the episode. The intuition behind our approach is to evaluate each trajectory by the potential cumulative reward it may gather past the planning horizon under the assumption that either one of the learned sampling policies is followed.\\

\textbf{Decentralized MPPI.} 
To obtain a coherent plan while preserving the complementary biases induced by $\pi_{\theta_s}$ and $\pi_{\theta_{bc}}$, we adopt a strategy analogous to Decentralized CEM~\cite{DecentCEMzhang2022a}, while retaining the exponential weighting used by MPPI. We maintain two independent MPPI proposal distributions over action sequences, one associated with each sampling policy.


At each MPPI iteration, each branch $i\in{s,bc}$ receives a budget of $N/2$ candidate trajectories. Of these, $N/2-n_{\mathrm{bias}}$ action sequences are sampled from the current MPPI proposal of that branch, while $n_{\mathrm{bias}}$ are generated using its corresponding learned sampling policy $\pi_{\theta_i}$. Candidate trajectories are simulated and evaluated using Eq.~\ref{eq:alg_evaluation}. The highest-scoring top-k candidates are then used to independently update each branch according to the exponential MPPI aggregation in Eq.~\ref{eq:mppi_update}.


After $K$ planning iterations, each branch yields a candidate plan given by the mean of its final action-sequence distribution. We evaluate both candidate plans using Eq.~\ref{eq:alg_evaluation} and execute the first action of the higher-scoring plan. Only the parameters of the selected planning distribution are stored in the replay buffer together with the resulting transition.


\subsection{Gated Planning Distillation}\label{sec:GPD}

This work learns policies that approximate the planner's conditioned action distribution, both to regularize $\pi_{\theta_s}$ and to train the cloned policy $\pi_{\theta_{bc}}$. Prior work pursuing similar objectives typically assumes access to the planner's current distribution. In our setting, satisfying this assumption would require re-planning from every sampled state, as the planning distribution depends on the current biasing distributions and bootstrap action-value functions. Recomputing these targets throughout training is therefore computationally prohibitive.

A cheaper alternative is to distill planning distributions stored in the replay buffer. However, these targets become stale as the policies and value functions evolve. Recent methods mitigate this issue through partial or periodic reanalysis~\cite{wang2025bootstrapped,schrittwieser2021online}, but doing so still incurs substantial computational overhead.

Rather than assuming access to an up-to-date planner, we study when stale planning targets remain useful for training the regularization policy $\pi_{\theta_{reg}}$ and the cloned sampling policy $\pi_{\theta_{bc}}$. Related work~\cite{zhan2025bootstrap} similarly recognizes that planning targets should not contribute equally, and reweights them using a normalized exponential function of their recorded episodic returns. We instead consider whether a planning target should contribute to distillation at all. To this end, we introduce \emph{Gated Planning Distillation}, which selectively distills stored planning distributions according to their value under the current learned policies.

Although preferentially distilling higher-value planning targets is intuitive, our gating criterion follows from the objectives governing the two sampling policies. Specifically, we show that the cloned sampling policy affects an upper bound on the planning objective, while the regularization policy affects an analogous bound on the KL-regularized policy objective. These results motivate gating stored planning targets according to whether their distillation is expected to improve the corresponding bound.\\

\textbf{Effects in planning: Cloned Sampling Policy.}
As shown in~\cite{it-mppi}, given an arbitrary sampling policy over action sequences that acts as prior distribution from which to sample trajectories, $\tau \coloneq (z_0, a_0, z_1, a_1,\ldots)$, with induced trajectory distribution $\hat\pi_s(\tau)$, the planning objective function is upper bounded by:

\begin{align}\label{eq:mppi_elbo}
    J(q)  = \mathbb{E}_{\tau\sim q(\tau)}[&R(\tau)] - \lambda \infdiv{q}{\hat\pi_s} \leq \\
    & \lambda \log\mathbb{E}_{\tau\sim\hat\pi_s}[\exp(\lambda^{-1}R(\tau))] \notag
\end{align}


Assuming candidate trajectories are generated from a two-component sampling scheme composed of the planning distribution $\pi_P$ and a biasing policy $\pi_i$, $i\in\{s,bc\}$, as commonly done in MPPI-based RL methods~\cite{hansen2024tdmpc2,lin2025td}. If a candidate is sampled from $\pi_P$ with probability $(1-p)$ and from $\pi_i$ with probability $p$, the induced trajectory distribution is $\hat\pi_s(\tau) = (1-p)\pi_P(\tau) + p\pi_i(\tau)$. 
Then,

\begin{align}\label{eq:mppi_upperbound}
    \mathbb{E}&_{\hat\pi_s}\left[\exp\left(\frac{1}{\lambda}R(\tau)\right)\right] =
    \\& (1-p)\mathbb{E}_{\pi_P}\left[\exp\left(\frac{1}{\lambda}R(\tau)\right)\right] + p\mathbb{E}_{\pi_i}\left[\exp\left(\frac{1}{\lambda}R(\tau)\right)\right]\notag,
\end{align}

where $R(\tau)$ is the cumulative reward gathered along the trajectory $\tau$. In practice $p$ is implemented as a fixed proportion of samples being taken from $\pi_i$ and $(1-p)$ from $\pi_P$. 

For fixed $p>0$, and holding the $\pi_P$ term fixed, the right-hand side of Eq.~\ref{eq:mppi_upperbound} is monotonically increasing in $\mathbb{E}_{\tau\sim\pi_i}[\exp(\lambda^{-1}R(\tau))]$. Since the logarithm is also monotonically increasing for $\lambda>0$, and the variational bound in Eq.~\ref{eq:mppi_elbo} is tight at the optimal $q$~\cite{it-mppi}, increasing this quantity raises the maximum achievable value of the KL-regularized objective in Eq.~\ref{eq:mppi_elbo}. Thus, a biasing policy with lower expected exponential return can impose a lower ceiling on the achievable objective of the planner, whereas improving this quantity relaxes that limitation.

Since $\pi_{\theta_{bc}}$ is trained by directly cloning planning distributions stored in the replay buffer, the preceding result motivates the first component of our gating mechanism. Each replay transition stores the action $a_P$ executed by the planner, corresponding to the mean of the resulting planning distribution, together with its covariance matrix. Rather than indiscriminately cloning every stored planning distribution through Eq.~\ref{eq:rkl_loss}, we only distill a replayed target when it is estimated to increase the expected exponential-return term in Eq.~\ref{eq:mppi_upperbound}, thereby avoiding targets estimated to lower the maximum achievable planning objective. 
Since computing the corresponding trajectory-level expectation exactly would require additional rollouts and is computationally expensive, we introduce the following tractable approximation:

\begin{align}\label{eq:bc_gating}
    \frac{1}{N_{bc}}\sum_{i=1}^{N_{bc}}\exp(\lambda^{-1}Q^{\pi_{\theta_{bc}}}(z,a_i)) \leq \exp(\lambda^{-1}Q^{\pi_{\theta_{bc}}}(z,a_{P})),
\end{align}

where $a_i\sim\pi_{\theta_{bc}}(\cdot|z)$. The left-hand side is a Monte Carlo estimate of the expected exponential action-value under the current cloned policy, whereas the right-hand side evaluates the same surrogate at the action taken by the planner at the transition. We use $Q^{\pi_{\theta_{bc}}}(z,a)$ as a tractable approximation of the expected return conditioned on $(z,a)$, replacing the trajectory-level exponential-return criterion with an action-value-based gating rule.\\


\textbf{Effects in planning: KL-regularized Sampling policy.}





A similar implication holds for the KL-regularized sampling policy. Here, stale planning targets do not affect the planning objective directly; instead, they shape the regularization policy $\pi_{\theta_{reg}}$, which determines the variational ceiling of the KL-regularized objective optimized by $\pi_{\theta_s}$. Consequently, distilling poor planning targets into $\pi_{\theta_{reg}}$ can reduce the maximum achievable value of the regularized policy objective.

Starting from the KL-regularized policy objective, we obtain an analogous variational bound:

\begin{align}\label{eq:KLRL_elbo}
    \mathbb{E}_{a\sim\pi_{\theta_s}}[Q^{\pi_{\theta_s}}(z,a)] &- \lambda KL[\pi_{\theta_s}|\pi_{\theta_{reg}}] \\
    &\leq \lambda \log\mathbb{E}_{\pi_{\theta_{reg}}}[\exp(\lambda^{-1}Q^{\pi_{\theta_s}}(z,a))]. \notag
\end{align}

\begin{figure*}[ht]
\centering
\includegraphics[width=\textwidth]{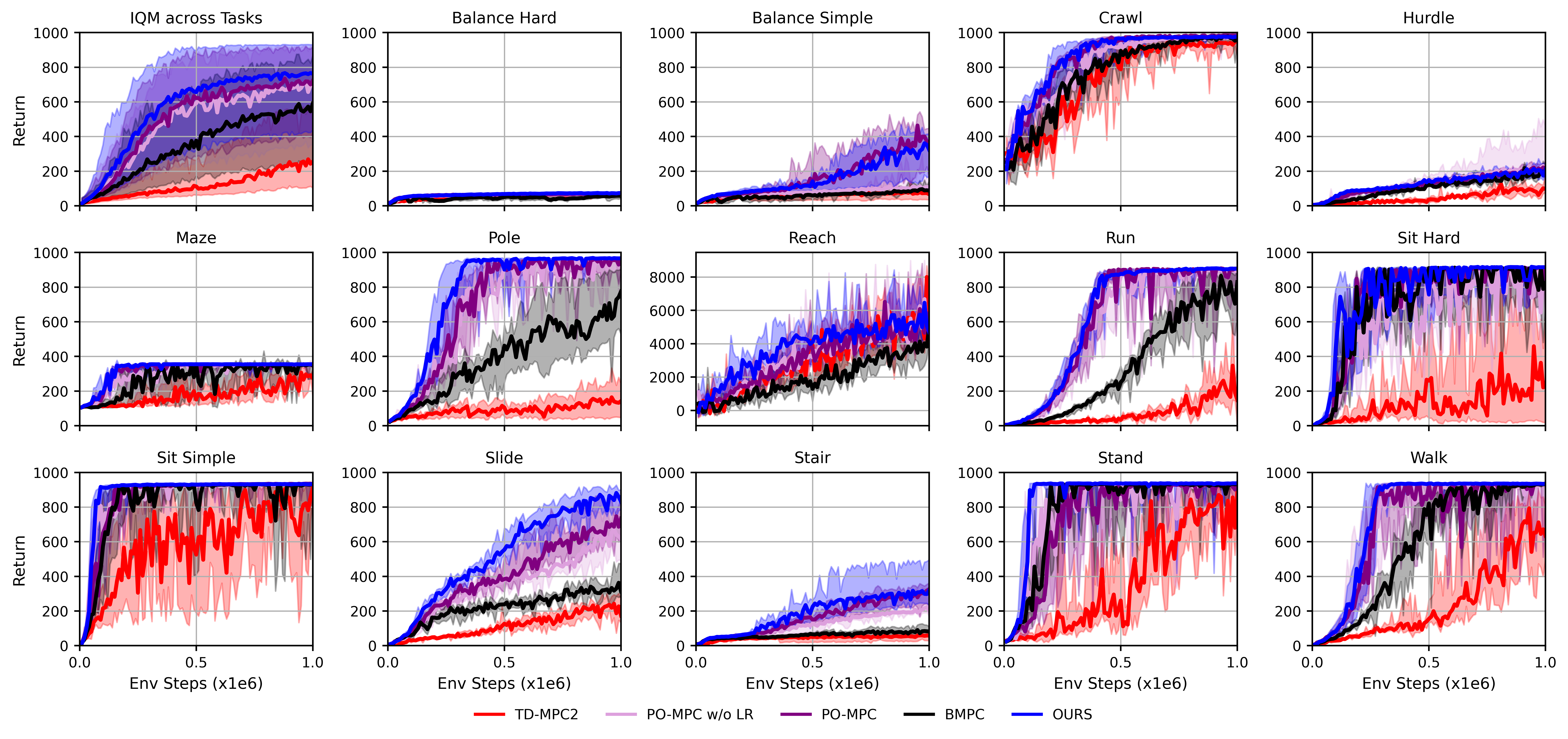}
\caption{Performance comparison in 14 state-based high-dimensional control tasks from HumanoidBench~\cite{humanoidbench}. Interquartile Mean (IQM) of 5 runs; shaded areas are the 95\% bootstrap Confidence Intervals (CI). In the top left, we visualize the Aggregate IQM with stratified bootstrap CI across all tasks except for \textit{Reach}, which has a different return range. GEM-MPC either matches or outperforms other baselines across tasks.}
\label{fig:main_results}
\end{figure*}

Thus, the maximum achievable value of the KL-regularized objective depends explicitly on the regularization policy $\pi_{\theta_{reg}}$.

For a fixed $Q^{\pi_{\theta_s}}$, increasing the right-hand side of Eq.~\ref{eq:KLRL_elbo} raises the maximum achievable value of the KL-regularized objective, although it does not necessarily imply an improvement in the resulting policy $\pi_{\theta_s}$. Conversely, decreasing this quantity imposes a lower ceiling on the objective optimized by $\pi_{\theta_s}$. We therefore apply an analogous gating mechanism as in Eq.~\ref{eq:bc_gating} when distilling the planning distribution into $\pi_{\theta_{reg}}$, retaining a replayed target only when its exponential action-value surrogate is at least as large as the corresponding expectation under the current regularization policy:

\begin{align}\label{eq:reg_gating}
    \frac{1}{N_{reg}}\sum_{i=1}^{N_{reg}}\exp(\lambda^{-1}Q^{\pi_{\theta_{s}}}(z,a_i)) \leq \exp(\lambda^{-1}Q^{\pi_{\theta_{s}}}(z,a_{P})),
\end{align}

where $a_i\sim\pi_{\theta_{reg}}(\cdot|s)$. Note that the action-value function in Eq.~\ref{eq:reg_gating} is that of the KL-regularized sampling policy, $Q^{\pi_{\theta_s}}$. Unlike Eq.~\ref{eq:mppi_elbo}, whose expectation is defined over complete trajectories, Eq.~\ref{eq:KLRL_elbo} is defined over single-step actions conditioned on the current state. This allows the exponential-value term to be evaluated directly using $Q^{\pi_{\theta_s}}(z,a)$, without introducing the trajectory-level return surrogate required for the cloned sampling policy.


\subsection{Method Summary.}
Our method addresses two limitations of MPPI-based reinforcement learning: the exploration-exploitation trade-off induced by relying on a single sampling policy during planning, and the instability caused by learning from stale planning distributions stored in replay. We address the first by augmenting MPPI with two complementary sampling policies: a KL-regularized policy $\pi_{\theta_s}$ that remains close to the planner while maximizing return, and a mode-seeking cloned policy $\pi_{\theta_{bc}}$ that captures the planner's most likely behavior. Together with their corresponding bootstrap action-value functions, these policies provide complementary exploratory and exploitative trajectory proposals, which are optimized independently through a decentralized MPPI procedure and compete to determine the executed plan. 

We address the second limitation through Gated Planning Distillation. Rather than indiscriminately cloning stored planning distributions or recomputing them through costly re-analysis, we update the cloned policy $\pi_{\theta_{bc}}$ and the planner proxy $\pi_{\theta_{reg}}$ only when a stored planning action improves the upper-bounds that depend on them (Eqs.~\ref{eq:mppi_elbo} and~\ref{eq:KLRL_elbo}). This Plan$\rightarrow$Evaluate$\rightarrow$Gate loop allows past planning experience to be reused selectively as the learned action-value functions evolve, retaining useful planner information while discarding stale targets that could restrict subsequent planning or policy performance. The resulting framework therefore combines complementary exploration and exploitation at planning time with a computationally efficient alternative to re-analysis for learning from replayed planning distributions.



\section{Experiments}
\label{sec:experiments}


\begin{table*}[t]
\centering
\caption{Ablations of \algname compared with PO-MPC~\cite{serragomez2026pompc}. Interquartile Mean (IQM) of 5 runs with 95\% bootstrap CI. 
}
\label{tab:method_comparison_summary}
\begin{tabular}{lccccc}
\toprule
\textbf{Method} & \textbf{Lazy Re-analyze} & \textbf{MoE} & \textbf{GPD} & \textbf{Aggregate IQM} & \textbf{Training Time (h)} \\
\midrule
PO-MPC w/o LR & \xmark & \xmark & \xmark & 697.0 [354.5, 918.5] & $6.4\pm 0.2$ \\
PO-MPC & \cmark  & \xmark & \xmark & 728.0 [431.3, 918.5] & $14.0\pm 0.8$ \\
\midrule
\algname (Ours) & \xmark  & \cmark & \cmark & \textbf{762.5 [436.2, 930.1]} & $9.2\pm 0.4$ \\
\midrule
\algname w/o GPD & \xmark  & \cmark & \xmark & \textbf{738.9 [391.6, 929.0]} & $8.3 \pm 0.3$ \\
 Only $\pi_{\theta_s}$ & \xmark  & \xmark & \cmark & \textbf{605.0 [273.8, 868.0]} & $7.0 \pm 0.2$ \\
Only $\pi_{\theta_{bc}}$ & \xmark  & \xmark & \cmark & \textbf{667.5 [315.8, 905.2]} & $6.8 \pm 0.2$ \\
\bottomrule
\end{tabular}
\end{table*}

\textbf{Experimental Setup.}
We evaluate different configurations of the proposed
framework (\algname) on the 14 continuous control tasks 
from HumanoidBench locomotion suite~\cite{humanoidbench}. These tasks are high-dimensional and cover a diverse range of continuous control challenges, including sparse reward, locomotion with high-dimensional state and action space ($\mathcal{A}\in\mathbb{R}^{61}$). All the experiments are run on a partition of an NVIDIA A100 GPU configured as a 4g.40GB Multi-Instance GPU (MIG), 
for 1e6 time-steps. In this work, we build upon the official JAX~\cite{jax2018github} implementation of PO-MPC~\cite{serragomez2026pompc}. As in PO-MPC, we inherit all architectural and pipeline choices from TD-MPC2, including data gathering through random walks for 1e4 total environment steps and the pre-training phase of the dynamic model and bootstrap action value functions. The architecture of all KL-regularized and bootstrap action value functions follow the same design. For reproducibility, our implementation and hyperparameters are available at~\githublink.\\

\textbf{Baselines.}
We empirically support the claims in this work by comparing \algname with other algorithms within the state of the art in MPPI-based RL, namely TD-MPC2~\cite{hansen2024tdmpc2}, BMPC~\cite{wang2025bootstrapped} and PO-MPC~\cite{serragomez2026pompc} with, and without, reanalyzing. Note that BMPC also uses Lazy Re-analyze. We also explore ablations of \algname by studying the effect of omitting any of the design choices introduced in Section~\ref{sec:method}. Table~\ref{tab:method_comparison_summary} provides an overview on the contributions of our work and their effects.

For our experiments, we employ the baseline implementations in JAX~\cite{serragomez2026pompc, flandermeyer2025bmpcjax}, which are either implemented by the original authors or in collaboration with them, since they reproduce the results from the original paper while increasing the computation speed. We evaluate our method under the same hyperparamers as PO-MPC, except for those related only to \algname. 


\subsection{Results}

The objective of this section is to test \algname from three perspectives. First, we show our method either surpasses or matches in performance the other baselines. Second, we make an empirical study over the effect of each of our contributions introduced in Section~\ref{sec:method}, comparing each ablation against PO-MPC, with and without Lazy Re-analyze, showing that their combination improves upon the state of the art both in terms of higher performance at lower training times. Finally, we further ablate our method, empirically analyzing each of the mechanisms introduced as part of each contribution.

Figure~\ref{fig:main_results} compares \algname against the baselines and illustrates the effect of Lazy Re-analyze in PO-MPC. Overall, combining Expert-augmented Planning (MoE) with Gated Planning Distillation (GPD) matches or improves upon the baselines across the high-dimensional HumanoidBench tasks, achieving the highest aggregate IQM of $762.5$, compared with $728.0$ for PO-MPC with Lazy Re-analyze. The largest gains are observed in \textit{Slide} and \textit{Stair}, where \algname reaches substantially higher returns, while its learning curves also exhibit greater stability in several tasks. Importantly, these gains do not require the computational cost of Lazy Re-analyze: \algname reduces training time from $14.0$ to $9.2$ hours while attaining higher aggregate performance. These results suggest that selectively exploiting useful past planning distributions provides a more favorable performance-compute trade-off than periodically recomputing them.

\subsection{Balancing Exploration and Exploitation}\label{sec:res_moe}


We investigate whether the performance gains of \algname arise from separating exploration and exploitation across sampling policies with distinct objectives. Starting from PO-MPC, where a single sampling policy must balance both behaviors through its regularization objective, we progressively disentangle these roles and evaluate different combinations of KL objectives: mode-seeking behavior for exploitation and support-covering behavior for broader exploration. Our results in Table~\ref{tab:method_comparison_summary} show that assigning these objectives to separate policies yields stronger and more stable performance, allowing the KL-regularized sampling policy to optimize over a broader action support while the cloned policy exploits high-probability planning behavior. Decentralized MPPI further improves robustness by maintaining separate planning proposals, consistent with improved robustness in tasks with multiple local optima, e.g. Sit Hard. 

\subsection{Effects of Gating Planning Data}\label{sec:res_GPD}
We next analyze the behavior of Gated Planning Distillation throughout training. Table~\ref{tab:method_comparison_summary} shows the impact of gating on performance. During training, the proportion of samples satisfying each of Eq.~\ref{eq:bc_gating} and~\ref{eq:reg_gating} is observed to drop rapidly below $50\%$ of the sampled replay data and steadily decline over training until reaching around $20\%$. Despite training on substantially fewer planning targets, the gated variant consistently achieves higher performance than indiscriminate distillation. This suggests that a large fraction of stored planning distributions becomes uninformative, or potentially detrimental, as the planner evolves, and that selectively discarding such targets can improve learning.

\section{Discussion and Conclusion}
\label{sec:discussion}
\textbf{Summary of Findings.}
Across 14 HumanoidBench tasks, \algname consistently improves upon the baselines. As shown in Table~\ref{tab:method_comparison_summary}, using both $\pi_{\theta_s}$ and $\pi_{\theta_{bc}}$ during planning already yields gains over PO-MPC, even without Gated Planning Distillation. Enabling GPD further widens this gap while remaining a more computationally efficient alternative to Lazy Re-analyze. Our mixture-of-experts ablations show that jointly using the exploratory and cloned sampling policies outperforms relying on either policy alone, supporting the role of complementary sampling objectives during planning. In turn, the GPD ablations show that selectively distilling replayed planning targets according to Eqs.~\ref{eq:bc_gating} and~\ref{eq:reg_gating} improves performance over indiscriminate cloning.
Together, these results support the two main design choices of \algname: combining complementary sampling policies improves the quality and robustness of planning, while Gated Planning Distillation provides a tractable mechanism for filtering stale planning targets and retaining those that remain useful for learning the cloned and regularization policies.\\

\textbf{Limitations and Future Work.}
Our first limitation concerns Gated Planning Distillation, which relies on learned action-value estimates to determine which replayed planning targets remain useful. Errors in these estimates may therefore affect the gating decisions, and our criteria should be interpreted as tractable surrogates rather than exact tests of target usefulness.

A second limitation is the additional memory and computation required relative to approaches using a single sampling policy, as our method maintains an additional policy and action-value function and evaluates trajectories under both. Finally, our method relies on Gaussian policy and planning distributions, which may be restrictive in high-dimensional or multimodal action spaces where high-value regions can exhibit more complex structure. More expressive policy classes, such as flow-matching policies that approximate Boltzmann distributions~\cite{zhong2026reparameterization}, provide a promising direction for representing richer sampling distributions and potentially reducing the need for multiple specialized components.

\addtolength{\textheight}{-12cm}   




\section*{ACKNOWLEDGMENT}
The authors used ChatGPT to assist generating code for plotting the results. The authors reviewed and verified the resulting code, and take full responsibility for the final manuscript, implementation, and reported results.


\bibliographystyle{IEEEtran}
\bibliography{IEEEabrv,references}

@inproceedings{hansen2024tdmpc2,
  title={{TD-MPC2: Scalable, Robust World Models for Continuous Control}}, 
  author={Nicklas Hansen and Hao Su and Xiaolong Wang},
  booktitle={International Conference on Learning Representations (ICLR)},
  year={2024}
}

@inproceedings{
wang2025bootstrapped,
title={{Bootstrapped Model Predictive Control}},
author={Yuhang Wang and Hanwei Guo and Sizhe Wang and Long Qian and Xuguang Lan},
booktitle={The Thirteenth International Conference on Learning Representations},
year={2025},
}

@article{lin2025td,
  title={{TD-M$\text{(PC)}^{2}$: Improving Temporal Difference MPC Through Policy Constraint}},
  author={Lin, Haotian and Wang, Pengcheng and Schneider, Jeff and Shi, Guanya},
  journal={arXiv preprint arXiv:2502.03550},
  year={2025}
}

@ARTICLE{it-mppi,
  author={Williams, Grady and Drews, Paul and Goldfain, Brian and Rehg, James M. and Theodorou, Evangelos A.},
  journal={IEEE Transactions on Robotics}, 
  title={{Information-Theoretic Model Predictive Control: Theory and Applications to Autonomous Driving}}, 
  year={2018},
  volume={34},
  number={6},
  pages={1603-1622},
  doi={10.1109/TRO.2018.2865891}}

@inproceedings{haarnoja2018soft,
  title={{Soft actor-critic: Off-policy maximum entropy deep reinforcement learning with a stochastic actor}},
  author={Haarnoja, Tuomas and Zhou, Aurick and Abbeel, Pieter and Levine, Sergey},
  booktitle={Int. conf. on machine learning},
  pages={1861--1870},
  year={2018},
  organization={Pmlr}
}

@article{moerland2023model,
author = {Moerland, Thomas M. and Broekens, Joost and Plaat, Aske and Jonker, Catholijn M.},
title = {Model-based Reinforcement Learning: A Survey},
year = {2023},
issue_date = {Jan 2023},
publisher = {Now Publishers Inc.},
address = {Hanover, MA, USA},
volume = {16},
number = {1},
issn = {1935-8237},
doi = {10.1561/2200000086},
journal = {Found. Trends Mach. Learn.},
month = jan,
pages = {1–118},
numpages = {130}
}

@article{tdmpc1,
  title={{Temporal Difference Learning for Model Predictive Control}},
  author={Hansen, Nicklas A and Su, Hao and Wang, Xiaolong},
  journal={International Conference on Machine Learning},
  pages={8387--8406},
  year={2022}
}

@misc{flandermeyer2025bmpcjax,
  author       = {Shane Flandermeyer},
  title        = {{bmpc-jax}: Jax/Flax Implementation of {BMPC}},
  howpublished = {\url{https://github.com/ShaneFlandermeyer/bmpc-jax}},
  year         = {2024},
  note         = {Accessed: 2025-08-28}
}

@software{jax2018github,
  author = {James Bradbury and Roy Frostig and Peter Hawkins and Matthew James Johnson and Chris Leary and Dougal Maclaurin and George Necula and Adam Paszke and Jake Vander{P}las and Skye Wanderman-{M}ilne and Qiao Zhang},
  title = {{JAX}: composable transformations of {P}ython+{N}um{P}y programs},
  version = {0.3.13},
  year = {2018},
}

@inproceedings{
schrittwieser2021online,
title={{Online and Offline Reinforcement Learning by Planning with a Learned Model}},
author={Julian Schrittwieser and Thomas K Hubert and Amol Mandhane and Mohammadamin Barekatain and Ioannis Antonoglou and David Silver},
booktitle={Advances in Neural Information Processing Systems},
editor={A. Beygelzimer and Y. Dauphin and P. Liang and J. Wortman Vaughan},
year={2021},
url={\url{https://openreview.net/forum?id=HKtsGW-lNbw}}
}

@INPROCEEDINGS{humanoidbench, 
    AUTHOR    = {Carmelo Sferrazza AND Dun-Ming Huang AND Xingyu Lin AND Youngwoon Lee AND Pieter Abbeel}, 
    TITLE     = {{HumanoidBench: Simulated Humanoid Benchmark for Whole-Body Locomotion and Manipulation}}, 
    BOOKTITLE = {Proceedings of Robotics: Science and Systems}, 
    YEAR      = {2024}, 
    ADDRESS   = {Delft, Netherlands}, 
    MONTH     = {July}, 
    DOI       = {10.15607/RSS.2024.XX.061} 
}

@inproceedings{bhardwaj2021blending,
title={{Blending {MPC} {\&} Value Function Approximation for Efficient Reinforcement Learning}},
author={Mohak Bhardwaj and Sanjiban Choudhury and Byron Boots},
booktitle={International Conference on Learning Representations},
year={2021},
}

@article{tirumala2022,
  author  = {Dhruva Tirumala and Alexandre Galashov and Hyeonwoo Noh and Leonard Hasenclever and Razvan Pascanu and Jonathan Schwarz and Guillaume Desjardins and Wojciech Marian Czarnecki and Arun Ahuja and Yee Whye Teh and Nicolas Heess},
  title   = {{Behavior Priors for Efficient Reinforcement Learning}},
  journal = {Journal of Machine Learning Research},
  year    = {2022},
  volume  = {23},
  number  = {221},
  pages   = {1--68},
  url     = {http://jmlr.org/papers/v23/20-1038.html}
}

@article{zhuang2025tdmpbc,
  title={{Tdmpbc: Self-imitative reinforcement learning for humanoid robot control}},
  author={Zhuang, Zifeng and Shi, Diyuan and Suo, Runze and He, Xiao and Zhang, Hongyin and Wang, Ting and Lyu, Shangke and Wang, Donglin},
  journal={arXiv preprint arXiv:2502.17322},
  year={2025}
}

@inproceedings{
zhan2025bootstrap,
title={{Bootstrap Off-policy with World Model}},
author={Guojian Zhan and Likun Wang and Xiangteng Zhang and Jiaxin Gao and Masayoshi Tomizuka and Shengbo Eben Li},
booktitle={The Thirty-ninth Annual Conference on Neural Information Processing Systems},
year={2025},
url={https://openreview.net/forum?id=zNqDCSokDR}
}

@inproceedings{
serragomez2026pompc,
title={A {KL}-regularization framework for learning to plan with adaptive priors},
author={Alvaro Serra-Gomez and Daniel Jarne Ornia and Dhruva Tirumala and Thomas M. Moerland},
booktitle={Forty-third International Conference on Machine Learning},
year={2026},
url={https://openreview.net/forum?id=zO8vzSGgTn}
}

@inproceedings{
DecentCEMzhang2022a,
title={A Simple Decentralized Cross-Entropy Method},
author={Zichen Zhang and Jun Jin and Martin Jagersand and Jun Luo and Dale Schuurmans},
booktitle={Advances in Neural Information Processing Systems},
editor={Alice H. Oh and Alekh Agarwal and Danielle Belgrave and Kyunghyun Cho},
year={2022},
url={https://openreview.net/forum?id=IQIY2LASzYx}
}

@inproceedings{
zhong2026reparameterization,
title={Reparameterization Flow Policy Optimization},
author={Hai Zhong and Zhuoran Li and Xun Wang and Longbo Huang},
booktitle={Forty-third International Conference on Machine Learning},
year={2026},
}

@article{evers2026efficienttdmpc,
  title={EfficientTDMPC: Improved MPC Objectives for Sample-Efficient Continuous Control},
  author={Evers, Thomas and Meo, Cristian and Bohmer, Wendelin and Dauwels, Justin and Oren, Yaniv},
  journal={arXiv preprint arXiv:2605.16692},
  year={2026}
}

@inproceedings{nair2018overcoming,
  title={Overcoming exploration in reinforcement learning with demonstrations},
  author={Nair, Ashvin and McGrew, Bob and Andrychowicz, Marcin and Zaremba, Wojciech and Abbeel, Pieter},
  booktitle={2018 IEEE international conference on robotics and automation (ICRA)},
  pages={6292--6299},
  year={2018},
  organization={IEEE}
}

@inproceedings{CRR_NEURIPS2020_588cb956,
 author = {Wang, Ziyu and Novikov, Alexander and Zolna, Konrad and Merel, Josh S and Springenberg, Jost Tobias and Reed, Scott E and Shahriari, Bobak and Siegel, Noah and Gulcehre, Caglar and Heess, Nicolas and de Freitas, Nando},
 booktitle = {Advances in Neural Information Processing Systems},
 editor = {H. Larochelle and M. Ranzato and R. Hadsell and M.F. Balcan and H. Lin},
 pages = {7768--7778},
 publisher = {Cur. Assoc., Inc.},
 title = {Critic Regularized Regression},
 volume = {33},
 year = {2020}
}

@inproceedings{
chang2026the,
title={The Surprising Difficulty of Search in Model-Based Reinforcement Learning},
author={Wei-Di Chang and Mikael Henaff and Brandon Amos and Gregory Dudek and Scott Fujimoto},
booktitle={Forty-third International Conference on Machine Learning},
year={2026},
}

\end{document}